\documentclass[10pt,twocolumn,letterpaper]{article}

\usepackage{wacv}
\usepackage{times}
\usepackage{epsfig}
\usepackage{graphicx}
\usepackage{amsmath}
\usepackage{amssymb}
\usepackage[normalem]{ulem}
\usepackage{enumitem}

\newcount\Comments
\usepackage{color}
\definecolor{darkgreen}{rgb}{0,0.6,0}
\definecolor{orange}{rgb}{1,0.5,0}
\newcommand{\kibitz}[2]{\ifnum\Comments=0{\color{#1}{#2}}\fi}
\newcommand{\zk}[1]{\kibitz{red}{[ZK: #1]}}
\newcommand{\yh}[1]{\kibitz{red}{[YH: #1]}}

\wacvfinalcopy 

\def\wacvPaperID{205} 

\ifwacvfinal\fi
\begin{document}

\title{Data-Efficient Graph Embedding Learning for PCB Component Detection}

\author{
Chia-Wen Kuo \\
Georgia Institute of Technology\\
{\tt\small albert.cwkuo@gatech.edu}
\and
Jacob D. Ashmore\\
Georgia Tech Research Institute\\
{\tt\small jacob.ashmore@gtri.gatech.edu}
\and
David Huggins\\
Georgia Tech Research Institute\\
{\tt\small david.huggins@gtri.gatech.edu}
\and
Zsolt Kira\\
Georgia Institute of Technology,\\Georgia Tech Research Institute\\
{\tt\small zkira@gatech.edu}
}

\maketitle
\ifwacvfinal\thispagestyle{empty}\fi

\begin{abstract}
This paper presents a challenging computer vision task, namely the detection of generic components on a PCB, and a novel set of deep-learning methods that are able to jointly leverage the appearance of individual components and the propagation of information across the structure of the board to accurately detect and identify various types of components on a PCB. Due to the expense of manual data labeling, a highly unbalanced distribution of component types, and significant domain shift across boards, most earlier attempts based on traditional image processing techniques fail to generalize well to PCB images with various quality, lighting conditions, etc. Newer object detection pipelines such as Faster R-CNN, on the other hand, require a large amount of labeled data, do not deal with domain shift, and do not leverage structure. To address these issues, we propose a three stage pipeline in which a class-agnostic region proposal network is followed by a low-shot similarity prediction classifier. In order to exploit the data dependency within a PCB, we design a novel Graph Network block to refine the component features conditioned on each PCB. To the best of our knowledge, this is one of the earliest attempts to train a deep learning based model for such tasks, and we demonstrate improvements over recent graph networks for this task. We also provide in-depth analysis and discussion for this challenging task, pointing to future research.
\end{abstract}

\begin{figure}[t]
\begin{center}
\includegraphics[width=1.0\linewidth]{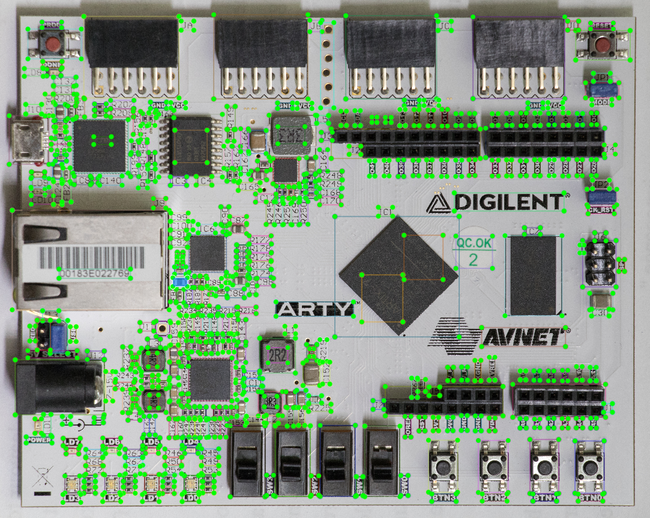}
\end{center}
\caption{Example of a labeled printed circuit board (PCB) image. The labeling process~\cite{lin2015label} is expensive due to the large amount of components and the expertise required to identify the correct type.}
%
%
\label{fig:label}
\end{figure}

\section{Introduction}
Detection of PCB components is an important step for automated PCB inspection and hardware reverse engineering. Traditionally, this step is carried out by a human expert to first locate and draw the bounding boxes of all components, and then identify the component types. However, it is extremely tedious and time consuming for a human expert to do such work. A PCB typically contains hundreds of components, and the identification of component types requires in-depth knowledge of electronic components. Figure.~\ref{fig:label} gives an example of a labeled PCB image. Based on our experience, it takes one to two days for a human expert to label a single PCB image. Thus, it would be extremely helpful if we can automate this tedious process.

Fine-tuning state-of-the-art object detection networks~\cite{girshick2015fast,ren2015faster} is an intriguing option for PCB component detection. However, it performs poorly due to following reasons:\yh{Original:However, directly fine-tuning current state-of-the-art object detection networks~\cite{girshick2015fast,ren2015faster} leads to poor performance on the task of PCB component detection. The task is more challenging due to several reasons:}

\textbf{Data Availability:} It is expensive to label a PCB image. This is also one of the objectives of this paper: to greatly reduce human efforts in labeling PCB components. Due to this expensive process, it is difficult to scale up a training dataset similar to other tasks (e.g. MSCOCO~\cite{lin2014microsoft} or PASCAL~\cite{everingham2010pascal}), especially as online data tends to have lower resolution than required or are taken from a side view, and hence manual image captures are required. For these same reasons, it is also difficult to obtain a large amount of unlabeled PCB images. Therefore, it is difficult to leverage large-scale unsupervised or semi-supervised learning methods to improve the accuracy.

\begin{figure}[t]
\minipage{0.57\linewidth}
  \includegraphics[width=\linewidth]{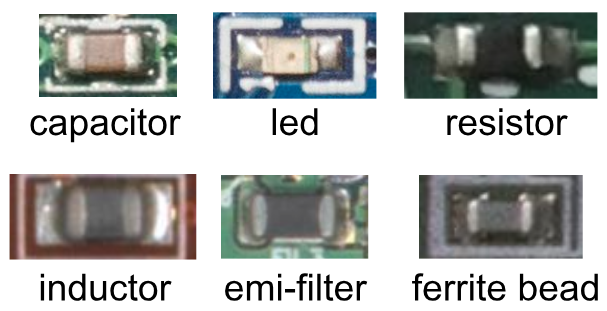}
\endminipage\hfill
\minipage{0.38\linewidth}
  \includegraphics[width=\linewidth]{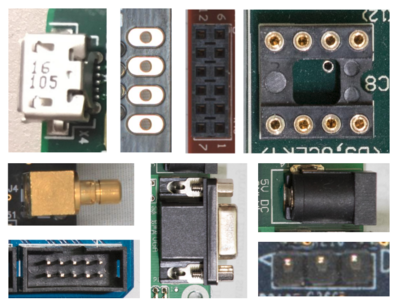}
\endminipage
\newline
\caption{\textit{Left:} low inter-class variance types. \textit{Right:} high intra-class variance type (connector).}
\label{fig:class_variance}
\end{figure}

\textbf{Data Distribution and Domain Shift:} The distribution of component types is highly unbalanced. For example, on a PCB we may have thousands of resistors and capacitors, while we may only have hundreds of integrated chips (IC) and only a few switches or some rare types such as an emi-filter or display. Please refer to supplementary materials for a more detailed figure of type distribution. Furthermore, the data distribution also exhibits high intra-class variance and low inter-class variance for certain types of components. In Figure.~\ref{fig:class_variance}, we show that resistors, capacitors, ferrite beads, etc. share similar visual appearance (low inter-class variance), while there are various kinds of connectors that all look very different (high intra-class variance).

Using hand-engineered features is another appealing strategy for a well-defined application. There are a few attempts for PCB component detection that are based on traditional image processing methods~\cite{li2013smd,li2014text,pithadiya2015evaluating,pramerdorfer2015pcb,pramerdorfer2015dataset,li2016localizing} before the deep learning era. However, many other factors such as image resolution, lighting condition, etc. limit the applicability of above methods. In this work we aim to address the problem of detecting a broad set of component types on a variety of PCBs. Furthermore, to the best of our knowledge, our proposed method is one of the earliest attempts to apply deep learning based method for such a task.\yh{Original:
There are many other factors such as the image resolution, lighting condition, etc. that make it even more ambiguous to correctly identify component types. Therefore, it is difficult to hand-engineer certain features to perform general-purpose PCB component detection task. There are a few attempts for PCB component detection that are based on traditional image processing or computer vision methods~\cite{li2013smd,li2014text,pithadiya2015evaluating,pramerdorfer2015pcb,pramerdorfer2015dataset,li2016localizing} before the deep learning era. However, most of them can only detect limited types of components or only work on easier data. In contrast, we propose the problem of detecting a broad set of component types on a variety of PCBs. Furthermore, to the best of our knowledge, our proposed method is one of the earliest attempts to apply deep learning based method for such a task.}





In this paper, we develop a dataset consisting of a small set of annotated boards across many different PCB types and varieties. To address the challenges arising from data availability and distribution, we propose a three stages object detection pipeline that combines traditional region-based networks ~\cite{girshick2015fast,ren2015faster} with low-shot similarity learning methods~\cite{koch2015siamese,vinyals2016matching,snell2017prototypical,garcia2017few}. Further, in order to fully exploit the structure inherent in the problem of PCB component detection, we design a new graph network block inspired by recent work in such networks. The key idea is that we are able to condition our classifier on the biases in how boards are laid out and propagation of a few labels on components on the test board. Our method therefore contains the following components: 

\textbf{Region Proposal Network (RPN):} In this stage, we train a class-agnostic RPN~\cite{ren2015faster} to propose potential component locations. By doing so, the detection network does not have to deal with the data issues mentioned earlier, but simply shift the burden to the next proposal classification stage, in which some low-shot classification methods can be applied more easily.

\textbf{Similarity Prediction Network (SPN):} If we have high-quality proposals from the first stage, then the second stage becomes a classification problem. However, it is still a nontrivial task due to the data issues. As a result, we borrow methods from low-shot learning to learn a good feature embedding for similarity prediction~\cite{koch2015siamese,vinyals2016matching,snell2017prototypical,garcia2017few}. The network in this stage will compare each proposal to some component templates cropped from the same test board when a few components are labeled (semi-supervised learning). The proposals are then classified as the most similar template.

\textbf{Graph Network (GN):} We also observe the fact that components on the same board tend to share similar visual appearance and that there is structure in how the boards are laid out. Therefore, we hypothesize that we can exploit such data dependence on each board to further improve classification accuracy. Thus, we propose a novel GN block~\cite{garcia2017few,defferrard2016convolutional,kipf2017semi,simonovsky2017dynamic,hamilton2017inductive,NonLocal2018,battaglia2018relational,yang2018glomo} to capture the data dependency and further refine the features. The proposed GN block helps the information propagate across nodes and edges on the graph representation of a board, and show improved results over existing non-local graph networks (NLNN)\cite{NonLocal2018}.

In sum, we make the following contributions:
\begin{enumerate}[topsep=0pt,itemsep=-1ex,partopsep=1ex,parsep=1ex,labelindent=0.0em,labelsep=0.2cm,leftmargin=*]
\item To the best of our knowledge, this is one of the earliest deep learning based approaches that succeeds in the task of PCB component detection. We provide a dataset of 32 off-the-shelf boards that are densely annotated with components and will release it publicly.\zk{Should we say 48 images of 32 boards? Does the same board appear twice (different images of it) or just front and back?}
%
%
%
%
%
\item We propose a pipeline consisting of similarity learning and a novel graph network block that outperforms existing non-local graph networks. The model is successfully trained with only a limited amount of data while achieving strong performance.
\item We show the potential of our proposed Graph Network to jointly solve various challenging problems in computer vision research, including detection with few training examples, skewed type distribution, domain bias, etc
\item In-depth analysis and discussion for this challenging task is provided, pointing out a path for future research on this topic.
\end{enumerate}

In the rest of this paper, we briefly review relevant works in Section 2. In Section 3, we propose our model and introduce each part in detail. In Section 4, we provide implementation details, and conduct extensive experiments and ablation study for our proposed model. We discuss the reasons for the failure cases in Section 5, and conclude with areas for future work in Section 6.

\section{Related Works}
\subsection{Object Detection}
The computer vision research community has made remarkable progress in object detection~\cite{girshick2015fast,ren2015faster} in recent years. Thanks to the reliable performance of these networks, they have greatly facilitated the development of self-driving cars, robotics, and many other practical applications. There are also many open-source implementations~\cite{huang2017speed} that are stable, easy to use, and well documented such that researchers can fine-tune their pre-trained models for task-specific applications.  However, almost all object detection networks need to be trained on a large-scale dataset for decent performance. As opposed to image classification, there are few attempts for low-shot learning in object detection. Unfortunately, in our task of PCB component detection, it is expensive to construct a large-scale dataset for fine-tuning such detection network. Furthermore, there is inherent ambiguity in classifying components due to intra-class variance. Therefore, we investigate methods that can leverage inherent structure in the data (namely, within PCB boards) which traditional detection methods do not do.

\subsection{Low-Shot Learning}
Compared to object detection, there is more research effort on low-shot learning for image classification. One promising branch of methods~\cite{koch2015siamese,vinyals2016matching,snell2017prototypical,garcia2017few} base their idea on learning a good feature embedding such that objects of the same type will cluster together in the learned embedding space. In our task, we borrow the idea from these methods combined with a novel graph network block to learn a graph embedding in a data-efficient way.


\subsection{Graph Network}
Although low-shot learning methods point out a promising direction for learning in a data-efficient way, it is not guaranteed to generalize across domains. We observe the fact that each board has different characteristics such as resolution, noise level, lighting condition, etc. In other words, there is domain shift across each PCB image, which is not properly addressed in low-shot learning methods. Meanwhile, we also observe that same-type components on the same board tend to look very similar, and that boards tend to be structured in particular ways (i.e. there is a bias in how board designers lay out components). We therefore propose that it would be beneficial to capture this data dependency within a board. This is where Graph Networks~\cite{defferrard2016convolutional,kipf2017semi,simonovsky2017dynamic,hamilton2017inductive,battaglia2018relational,yang2018glomo} come in to play.

Graph Networks have become a popular research topic recently (see~\cite{battaglia2018relational} for an overview of various graph networks). In fact, the representation and optimization on graph models have existed for a long time, but it was not until several works proposed an efficient approximation to the original formulation that really sparked the research interest in Graph Networks~\cite{defferrard2016convolutional,kipf2017semi}. In the latest Graph Networks paper~\cite{battaglia2018relational}, they propose a generalized GN block that can be easily plugged in any network just like other standard building blocks for deep learning such as a convolutional or fully connected layer. In that paper\yh{need reference}, they advocate the power of a GN block to capture the structure and dependency between nodes and incorporating relational inductive biases into the graph representation. Inspired by this work, we design a novel GN block for the PCB detection task that takes into account node, edge, and global board embeddings jointly, and is more effective than other recent blocks such as the non-local graph network (NLNN)~\cite{NonLocal2018}.\zk{We may want to cite scene graph work, and mention they deal with message passing/structure but typically deal with many fewer components. That's for a different task though, so not completely necessary if there is no room}  By capturing these structural dependency, our work shows the potential of Graph Network to jointly solve various challenging problems in computer vision research, including detection with few training examples, skewed type distribution, domain bias, etc.

It's also worth pointing out that although the formulation of our GN module is similar to the NLNN, which was concurrently with our development of this paper, it was targeted to a completely different task. It aims to pass information between pixels on a feature map within a single image or a sequence of video image frames. Their aim is to capture the long-term dependency in terms of space or time, which can be interpreted as a way to adaptively increase receptive field.

Compared to previous Graph Network research, which mostly relies on a pre-defined graph structure, our proposed GN module, to the best of our knowledge, is one of the earliest works that is capable of jointly optimizing both the graph connectivity matrix (by backpropagating through the similarity predictions used to create them) and the feature embeddings (another most relevant work \cite{garcia2017few}). Similar ideas can be found in a series of Scene Graph works \cite{chen2018iterative, li2018factorizable, liu2018structure, xu2017scene}. Differing from our work, those are targeted to generate the triplet relationship of subject-predicate-object by iteratively passing information between nodes (detected objects) on an image. However, those methods do not scale well to an image with hundreds of objects, which is very common in our task.


\begin{figure}[t]
\begin{center}
\includegraphics[width=1.0\linewidth]{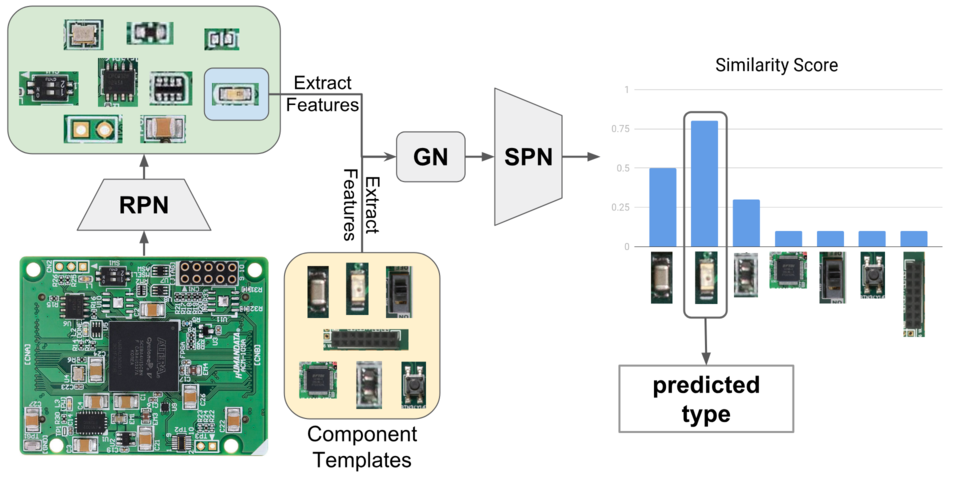}
\end{center}
\caption{Proposed three-stage pipeline. The RPN will take in a testing PCB image and output object proposals (green block). The Graph Network (GN) refines features based on structural information, and is jointly optimized with the similarity prediction network (SPN) which will then compute the similarity score between each proposal (blue block for example) and all component templates (yellow block). The proposal is then classified as the most similar template.}
\label{fig:three_stage_pipeline}
\end{figure}

\section{Model}
We propose a three-stage pipeline to address the challenges of PCB component detection as shown in Figure.~\ref{fig:three_stage_pipeline}. The first stage is a class-agnostic RPN which takes in a PCB image and propose bounding boxes for candidate components. The second stage is a similarity prediction network tailored to address the aforementioned data issues. The inputs to the second stage are object proposals. It will then predict the type of these proposals by computing the similarity score against some template components, which is based on the idea from a branch of low-shot image classification methods~\cite{koch2015siamese,vinyals2016matching,snell2017prototypical,garcia2017few}. Finally, we develop a novel GN block to incorporate structure within a PCB to deal with issues such as high intra-class variation, and optimize refined features both for each node (component) as well as an embedding representing the entire board.

\subsection{Region Proposal Network}
The proposed pipeline is similar to~\cite{ren2015faster} in that the first stage is a class-agnostic RPN and the second stage is a classifier. The class-agnostic RPN is very helpful in dealing with the unbalanced type distribution. For certain types that only have tens or even only few examples in the dataset, it is difficult to directly predict their types. However, based on the idea that humans do not need to know the type but can still locate and draw the bounding box around the components, we shift the burden of low-shot classification to the latter stage. By aggregating all types of components into one single \textit{component} type, we now have enough data to train a good RPN which achieves high recall and precision.

\begin{figure}[t]
\begin{center}
\includegraphics[width=1.0\linewidth]{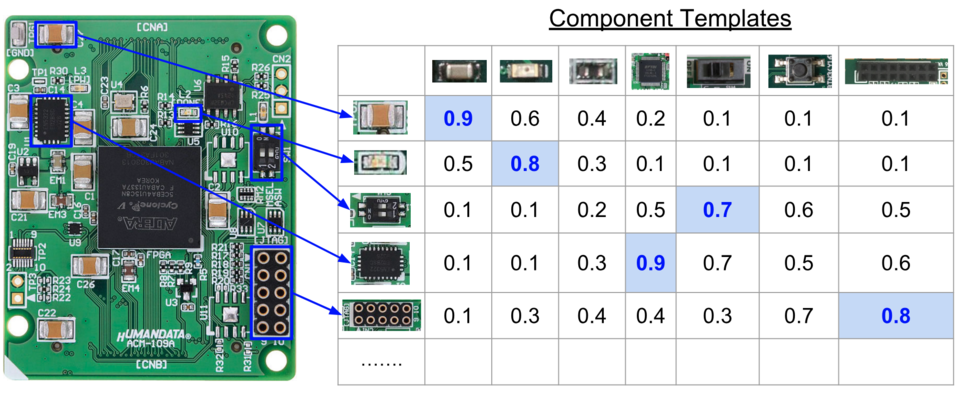}
\end{center}
\caption{The SPN will compute the similarity score between all pairs of proposal and component template. The proposals are then classified as the most similar template}
\label{fig:similarity_prediction}
\end{figure}

\subsection{Similarity Prediction Network (SPN)}
In this stage, we develop a method to classify each proposal from previous stage. A standard approach is to train an image classifier that directly outputs the probability distribution of types. However, due to the type imbalance and data deficiency issues, the trained classifier may simply overfit to the training set. Thus, we borrow ideas from low-shot learning and train a similarity prediction network. The SPN learns to predict a pairwise similarity score between each pair of training data. Therefore, if we only have $N$ components in the training set, by training with pairwise similarity, we actually have $N^2$ pairs of training data. This can be regarded as a way of data augmentation, which greatly helps with data sparsity issues.

The SPN is trained with triplet loss~\cite{hoffer2015deep,schroff2015facenet,hermans2017defense} instead of a na\"ive binary cross-entropy (BCE) loss, which predicts 1 if the pair represents the same type and 0 if it represents different types. As mentioned earlier in this paper, we observe that the PCB dataset exhibits the characteristics of low inter-class variance for certain types, e.g. inductor, resistor, and ferrite bead. Therefore, the BCE loss may overly penalize these types of components that share close visual similarity. Triplet loss is preferable here to only capture the relative similarity. Specifically, we want the similarity score between two resistors higher than the score between a resistor and a capacitor. Nevertheless, we don't want to drive the score to zero for the resistor-capacitor pair since they actually look similar. Thus, in a batch of training samples, we compute the triplet loss as:
\begin{equation}
\mathcal{L} = \frac{1}{M}\sum_{i=1}^{B}\sum_{s=1}^{S}\sum_{d=1}^{D} hinge( d(\phi_{x_i}, \phi_{x_d}), d(\phi_{x_i}, \phi_{x_s}))
\end{equation}
, where
\begin{equation}
hinge(d_1, d_2) = max(0, d_1 - d_2 + margin)
\end{equation}
\begin{equation}
d(\phi_{x_i}, \phi_{x_j}) = distance(\varphi_d(\phi_{x_i}), \varphi_d(\phi_{x_j}))
\end{equation}
$\phi_{x_i}$ is the feature vector of image $i$ in a training batch. $hinge(d_1, d_2)$ is the hinge loss with a hyper-parameter $margin$. $d(\phi_{x_i}, \phi_{x_j})$ can be any similarity function between feature $\phi_{x_i}$ and $\phi_{x_j}$ in some embedding space computed by arbitrary embedding function $\varphi_d$. Cosine similarity, for example, is a common choice of a similarity function. $s$ is the index of images in the batch that share the same type with image $i$, while $d$ is the index of images that are of a different type. $B$ is the batch size, while $S$ and $D$ refer to the number of similar and dissimilar pairs compared with image $i$, respectively. The total loss is normalized with $M$, the number of \textit{non-zero} terms of the hinge loss. This was shown in~\cite{hermans2017defense} to be a simple and effective way of hard negative mining for training a triplet loss.

By training the SPN with triplet loss, we pull the features of similar components close together while push those dissimilar away from each other. During testing, we simply compute the similarity scores between the object proposal and a set of component templates, and assign the type of the component template with the highest similarity score to the proposal. This process is illustrated in Figure.~\ref{fig:similarity_prediction}, and forms our SPN baseline before incorporating structure within PCBs using graph networks.

\subsection{Graph Network}
To further improve the accuracy of the SPN, we apply a GN block to refine the features. We observe the fact that same-type components tend to look similar within the same PCB but dissimilar across different PCBs. In other words, there is a domain shift with a bias across each PCB and we leverage this bias in a semi-supervised fashion where a few components on the board have been provided labels. Given this, there is also structure that can be leveraged due to patterns in how boards are laid out; for example, certain components may frequently appear with each other on boards. To leverage this structure, we apply a GN block to capture the data dependency within the same board. In order to fully capture this idea, we expand upon related work in graph networks~\cite{NonLocal2018} by adding global board-level features in addition to node and edge-level features \cite{battaglia2018relational}. This can be viewed as a feature refinement process conditioned on the data dependency between components on the same board as well as global board features.

In the Graph Network formulation, each node contains the feature representation of an object proposal computed from some feature extractor network. Different from the typical GN approach where the structure is known (e.g. for social media graphs), we use a learnable similarity (Eq. \ref{eq:fe} and \ref{eq:edge})\yh{where is its description? eq6?} to predict the edge feature between each pair of nodes, and thus construct a dense graph. The GN block propagates node information along the edges.

\begin{figure}[t]
\minipage{0.43\linewidth}
  \includegraphics[width=\linewidth]{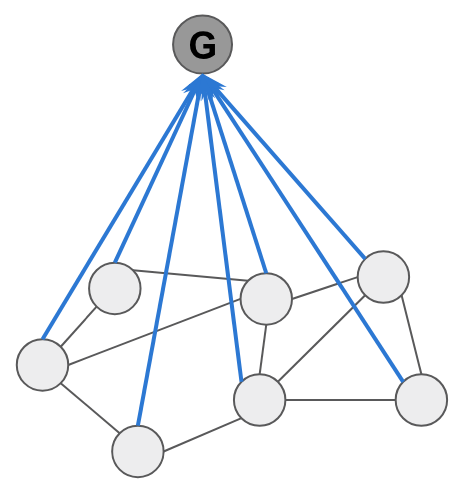}
\endminipage\hfill
\minipage{0.53\linewidth}
  \includegraphics[width=\linewidth]{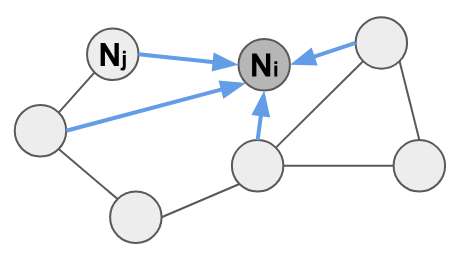}
\endminipage
\newline
\caption{\textit{Left:} Aggregate node features to compute the global graph feature. \textit{Right:} Aggregate edge features for node $i$.}
\label{fig:gn_block}
\end{figure}

Here we describe our novel GN block architecture:

\textbf{Global Feature:} Unlike the NLNN, we include a global embedding for the entire scene (board) and incorporate it later in the node embedding. The GN block therefore first computes a global feature $\phi_g$ of the whole graph (board) by aggregating the features of all nodes in the graph:
\begin{equation}
\phi_g = mean(\{\varphi_g(\phi_n)\})
\end{equation}
, where $\varphi_g$ is an arbitrary embedding function (e.g. fully connected layer). $\{\varphi_g(\phi_n)\}$ denotes the set of all embedded node features, $\varphi_g(\phi_n)$, in the graph. This process is illustrated in Figure.~\ref{fig:gn_block} Left.

\textbf{Edge Features:} The second step is to compute the edge features, aggregate them, and pass them to the associated nodes, as illustrated in Figure.~\ref{fig:gn_block} Right. We adopt the idea from~\cite{NonLocal2018,yang2018glomo} to compute and aggregate edge features. Specifically, for each node $i$, the aggregated edge feature $\phi_{e_i}$ is computed as:
\begin{equation}\label{eq:fe}
\phi_{e_i} = \sum_j w_{i,j}\varphi_e(\phi_{n_j})
\end{equation}
, where $w_{i,j}$ is the normalized dot product similarity. Specifically, it first computes the unnormalized dot product similarity $\hat{w}_{i,j}$ by:
\begin{equation}\label{eq:edge}
\hat{w}_{i,j} = dot(\psi_1(\phi_{n_i}), \psi_2(\phi_{n_j}))
\end{equation}
, and then normalize $\{\hat{w}_{i,j}\}_{j=1...n}$ by squared ReLU:
\begin{equation}
w_{i,j} = \frac{ReLU^2(\hat{w}_{i,j})}{\sum_{j}ReLU^2(\hat{w}_{i,j})}
\end{equation}
According to~\cite{yang2018glomo}, normalization using squared ReLU gives us a sparse graph, which is preferable as opposed to Softmax, which gives us a densely connected graph.

$\varphi_e$, $\psi_1$, and $\psi_2$ are arbitrary embedding functions.

\textbf{Node Features:} Differing again from NLNN, we then pass the global feature $\phi_g$ and the aggregated edge feature $\phi_{e_i}$ to node $i$. The final node feature $\hat{\phi}_{n_i}$ is computed as:
\begin{equation}
\hat{\phi}_{n_i} = ReLU(\phi_{n_i}+\varphi_n(concat(\phi_{n_i}, \phi_{e_i}, \phi_g)))
\end{equation}
, where $\varphi_n$ is another embedding function.

\yh{Above function looks like a recursive function since \phi_{n_i} occurred in both left and right side. Try \hat{\phi_{n_i}} for the right ones.}\zk{Add information about how labels are concatenated with node features}There are five learnable embedding functions: $\varphi_g$, $\varphi_e$, $\varphi_n$, $\psi_1$, and $\psi_2$ in the GN block. The GN block can be plugged in between the feature extractor and the similarity predictor. It works just like a fully connected layer in terms of I/O format, which can be jointly optimized with the feature extractor and SPN using the triplet loss. We show in experiments below that this joint optimization of features, similarity, and structure outperforms similarity-only and NLNN. Furthermore, extra features, such as geometry or label info, can be concatenated with visual features before sending into the GN block. We show in experiments below that by incorporating label info, we can further improve the classification accuracy.

\section{Experiment}
To test our proposed model, we've created a PCB dataset which contains 48 images in total\yh{Duplicate? (we will release this dataset upon acceptance)}. Although the proposed model is extremely flexible, we instantiate one for the experiments and provide implementation details. Furthermore, a series of experiments and ablation studies are conducted to show the effectiveness of each design choice.

\subsection{Dataset}
We collected and labeled 48 PCB images from human experts to construct a dataset for training and testing. It is split into 40 for training and 8 for testing. Some of the PCB images are from the Internet while others are taken from a DSLR camera or an industrial camera. Consequently, the lighting condition, resolution, quality. etc. vary substantially across images. Nonetheless, our proposed model generalizes well and succeeds in detecting and classifying the components. As the labeling process is tedious and takes a few days for a human expert to complete a single board, it is expensive to scale up the dataset, encouraging us to seek a data-efficient approach. More dataset details can be found in the supplementary materials.


\subsection{Implementation Detail}
\zk{We might be able to put some of these in supplemental}
\textbf{Network architecture:} The RPN is fine-tuned from the ResNet-50 Faster R-CNN model pre-trained on COCO. In order to capture components of dramatic aspect ratio and scale variance, we use anchors of scales: $\{\frac{1}{32}, \frac{1}{16}, \frac{1}{8}, \frac{1}{4}, \frac{1}{2}, 1, 2, 4, 8\}$ and aspect ratios: $\{\frac{1}{16}, \frac{1}{8}, \frac{1}{4}, \frac{1}{2}, 1, 2, 4, 8, 16\}$\zk{is this still true?}. The model is trained for 90,000 steps with initial learning rate set to 3e-3 and reduced by a factor of 10 at step 40,000 and 70,000. We use aggressive data augmentation strategies including vertical and horizontal flipping, $90^\circ$ rotation, random cropping, as well as brightness, contrast, hue, saturation, and color jittering. We use the stable Faster R-CNN implementation from TensorFlow~\cite{huang2017speed} and keep other hyper-parameters their default values.

The SPN can either use the RoI pooled features from the RPN or train a standalone feature extractor to extract the feature of each proposal. For the convenience of the ablation study, we fine-tune a lightweight ResNet-18~\cite{he2016deep} model pre-trained on ImageNet~\cite{ILSVRC15} as a feature extractor. The dimension of the output feature is $d=512$. \zk{we should try resnet-50. BTW isn't this handicapping it compared to FRCNN which used Resnet50?}

We chose dot product as the similarity function to train the triplet loss. The embedding function $\varphi_d$ is a simple fully connected layer $fc(d, d/2)$ that maps the original $d$ dimensional feature vector to a lower dimensional feature embedding. The final accuracy is not sensitive to the \textit{margin} parameter in triplet loss. It is simply set to 1.0 for all experiments.

For the GN block, all five embedding functions are implemented by a simple fully connected layer. $\varphi_g$, $\varphi_e$, $\psi_1$, and $\psi_2$ are $fc(d, d/2)$, while $\varphi_n$ is $fc(2d, d)$ simply to match the dimensions. We append one GN block after the feature extractor.

The SPN, which includes a feature extractor followed by a GN block, is jointly optimized by triplet loss. We use SGD~\cite{bottou2010large} optimizer with learning rate 1e-4, momentum 0.9 and weight decay 1e-4. The learning rate is reduced by a factor of 2 if the evaluation accuracy does not improve over 50 epochs. In total, we train the SPN for 500 epochs. 

\textbf{Data Split:} Due to the limited amount of labeled data we have in the PCB dataset, it is unfeasible to split the data into train-val-test sets. Therefore, we use three-fold cross validation. However, differing from typical cross-validation, we want to make sure the types of components in the test set is the subset of the types in the training set. Therefore, we generate three different split configuration files. Each file contains a list of 33 training data and 15 testing data. To meet the criteria, we'll first sample the PCB images such that all types are covered at least once in the training set. The reset of the images can then be randomly sampled safely.

\textbf{Data Batching:} To train the SPN, we adopt the standard $N$-way $K$-shot data batching scheme for training a low-shot image classification network~\cite{vinyals2016matching} \zk{citation?}. Meanwhile, we also want to capture the data dependency on each PCB by the GN module. Therefore, for each iteration, we randomly sample one PCB image, and for that image, sample $N=10$ types of component and $K=10$ instances for each type of component. If there are less than $N$ types or $K$ instances on the board, we simply sample with replacement. In each instance, we will go through a random jittering process, such that the duplicated instances will still be different. We also perform experiments where multiple boards are sampled for each batch, although it did not perform as well.
We adopt an aggressive data augmentation strategy to address the data deficiency issue. The components are first randomly rotated by one of $\{ 0^\circ, 45^\circ, 90^\circ,..., 315^\circ \}$. Then their color, including brightness, contrast and saturation, is randomly jittered. We also randomly crop, resize, and flip the image patches.

\textbf{Testing:} During testing, we need to generate a set of component templates to which the SPN compares the cropped unknown components. We have tried several methods to extract the template(s) for each type: 1) randomly select one instance per type on the same board as template (in the semi-supervised setting), 2) for each type of components, compute the nearest neighbor in training to the centroid of the feature vectors, and 3) compute Silhouette Coefficient to determine the number of clusters and use k-means to computer cluster centers over the training. Since our method simply determines the type by the predicted similarity scores, it is a natural extension to include multiple templates. However, although we do not show this in the experimental results, by including multiple templates the performance actually decreases. This is likely because of the ambiguity resulted from low inter-class variability.

To run the whole pipeline, since the resolution is high, we divide the image into four parts, run the RPN on each part, and then merge the results. We threshold the output proposals from RPN at 0.3\zk{Do you need a threshold for mAP?}. To determine the types of the object proposals, we compute the similarity score between each proposal and template. The proposals are then classified as the most similar template.

\subsection{Ablation Study}
 
Our results include comparison with {\bf 1)} direct fine-tuning of a Faster R-CNN model pre-trained on MSCOCO (FRCNN), {\bf 2)} replacing the similarity prediction network with a classifier (CLF), {\bf 3)} use of a batching scheme during training that includes only components within the same board (within) or components across boards (across) where we sample $N=10$ types of components and $K=10$ instances per type randomly selected across PCBs at each iteration, {\bf 4)} training the SPN with a BCE or triplet loss (Loss column), {\bf 5)} use of no graph block, NLNN graph block, or our GN block, and {\bf 6)} inclusion of labels or geometry features, such as size, aspect ratio, location, etc, for the instance templates within the node embedding.
For each condition, we show classification accuracy (Top1/Top5) when using the ground truth proposals and mean average precision (mAP) of the whole pipeline including both detection and classification.

\begin{table*}
\begin{center}
\begin{tabular}{|l|c c c c|c c|}
\hline
Method & Batching & Loss & GN Block & Extra Features & CLF Top1 Acc & Pipeline mAP \\
\hline\hline
Baseline FR-CNN & - & - & - & - & - & 0.475 \\
CLF & within & cross entropy & $\times$ & - & 0.709 & 0.503 \\
CLF-GN & within & cross entropy & ours & $\times$ & \textbf{0.715} & \textbf{0.509} \\
\hline
SPN-B-A & across & BCE & $\times$ & $\times$ & 0.352 & 0.361 \\
SPN-T-A & across & triplet & $\times$ & $\times$ & 0.702 & 0.555 \\
SPN-T-A-GN & across & triplet & ours & $\times$ & 0.675 & 0.543 \\
SPN-T-W-NLNN & within & triplet & NLNN & $\times$ & 0.793 & 0.611 \\
SPN-T-W-GN & within & triplet & ours & $\times$ & \textbf{0.803} & \textbf{0.629} \\
\hline
SPN-T-W-GN-GF & within & triplet & ours & geometry & 0.816 & 0.639 \\
SPN-T-W-GN-LF & within & triplet & ours & label & \textbf{0.820} & \textbf{0.653} \\
\hline
\end{tabular}
\end{center}
\caption{Quantitative results for the ablation study. We test for different settings, and for the classification accuracy with ground-truth bounding boxes (second to the last column) as well as the whole pipeline using bounding boxes predicted from the RPN (last column).\zk{Additional expts/results: 1) t-sne, 2) precision/recall, 3) qualitative results, 4) Include labels, 5) Add to this table using just external templates, 6) test what happens when you add spatial features, 7) show that RPN is a bottleneck by training/testing with GT ROIs}}
\label{tab:ablation}
\end{table*}


\subsection{Results}

\begin{figure*}
\begin{center}
\includegraphics[width=0.95\linewidth]{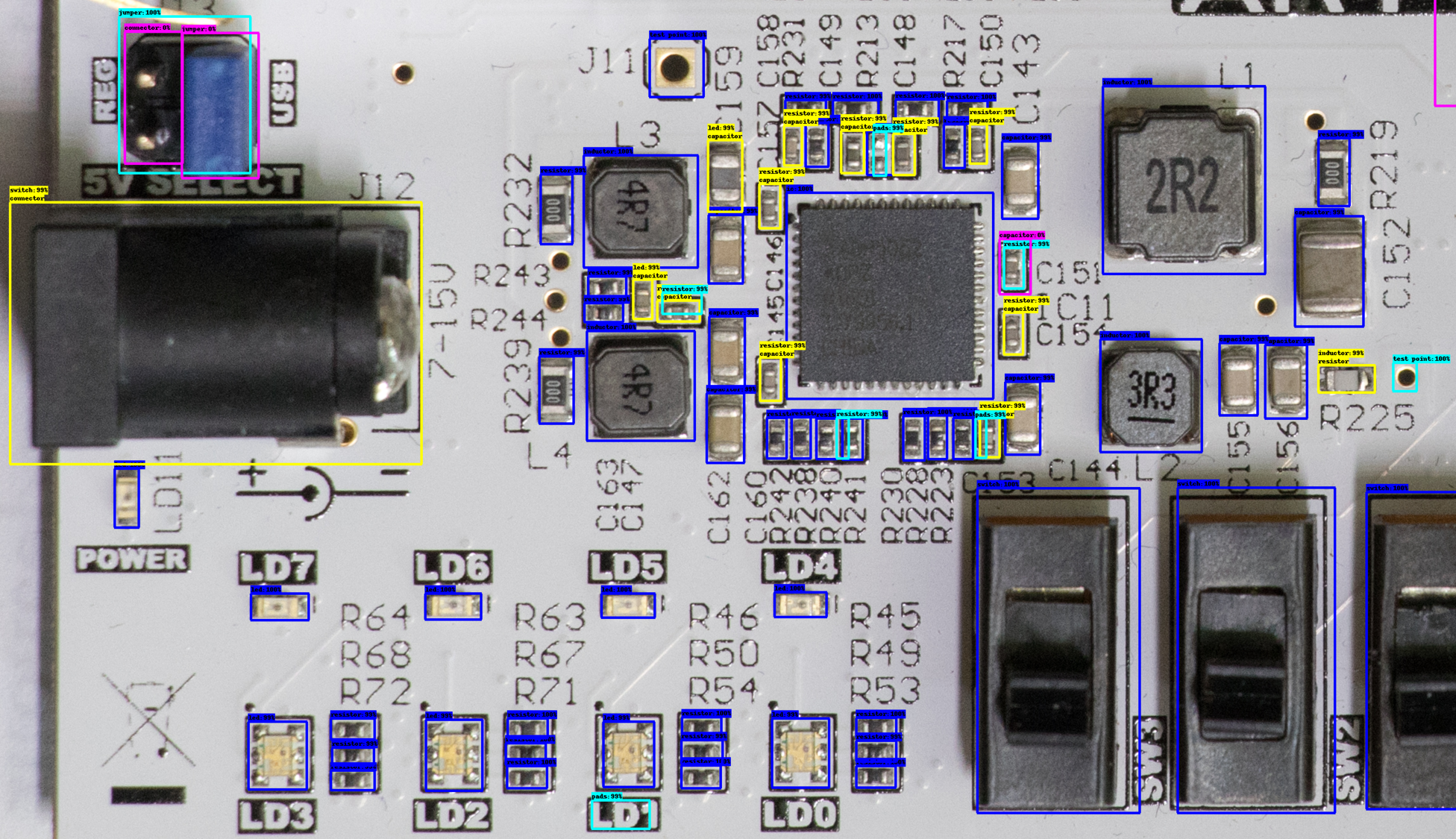}
\end{center}
\caption{Cropped detection result of the whole pipeline that use bounding boxes predicted from the RPN and then classify the proposals using proposed SPN. \textit{Blue:} correct type and precise bounding box location; \textit{Cyan:} imprecise bounding box location; \textit{Magenta:} miss-detected component; \textit{Yellow} precise bounding box location but wrong type. A higher resolution figure is presented in the Supplementary Materials. \zk{what does this mean? Are there classifications? We have to think about how to present these images, since it's hard to see anything unless you zoom in (I am about to try that anyways)}}
\label{fig:det}
\end{figure*}

Table \ref{tab:ablation} shows the results of our ablation study averaged across three-fold cross-validation. Our best model (SPN-T-W-GN-LF) is an SPN with features refined using our GN, which incorporates instances within the test board, and trains a pairwise similarity score using the triplet loss. Overall, this validates our hypothesis that {\it jointly} leveraging low-shot learning methods, some labeled instances from the board (and including the labels in the node representation to propagate them), and inherent structure in PCBs are important to achieve high accuracy on this challenging task. 

The fine-tuned Faster R-CNN (FR-CNN) and classification baselines (CLF-X) are able to achieve a mAP of approximately 0.5. Note that these conditions do not use labeled instances from the test boards, so these results cannot be directly compared with the rest. However, the low-shot methods using SPN (SPN-B-A and SPN-T-A) use on-board templates and hence form our baseline condition. The triplet loss is superior to BCE, which was not able to properly converge due to the difficulty of the training in terms of inter and intra-class variances discussed earlier. 
 
 We then show that adding our GN block (X-GN) improves performance over the triplet similarity baseline, and that batching within boards is better than across boards (SPN-T-A-GN at 0.543 mAP versus SPN-T-W-GN at 0.629 mAP). This is likely because the labeled templates come from the board during testing and allow us to deal with domain shift bias. Note that the improvement of GN is higher when compared to NLNN~\cite{NonLocal2018} (SPN-T-W-GN at 0.629 mAP versus SPN-T-W-NLNN at 0.611 mAP) showing that our Graph Network block is superior to other blocks since it aggregates a global feature representation, hence combining node, edge, and global features in a novel way. Interestingly, adding the label information directly into the node embedding, such that it is propagated (rather than simply used during the similarity-based matching to estimate the class label) produces even more improvements (SPN-T-W-GN-LF at 0.653mAP). Adding geometric information such as component positions and sizes also improves results, but less than labels. 


\section{Discussion}
As we examine the failure cases for classification, we observe that some result from the high intra-class and low inter-class variance issues that we mentioned in the Introduction section. Other causes for failure can be the ambiguously and subjectively defined component types and bounding boxes. For instance, the test point components are sometimes in the form of pins, pads or connectors. They are labeled as test points because of their definition in the data sheet or the silk screen label around them. Additionally, it is hard to define the bounding boxes for some pins and connectors. Should they be grouped by their function on the PCB or should each individual pin be labeled as a new instance of a pin? There are some more extreme cases such as the resistor jumper and capacitor jumper, which are actually compositions of a resistor/capacitor and a pad. Should we consider it as a single instance of a resistor/capacitor jumper or separately as its individual components? In order to resolve these ambiguous cases, some external knowledge should be incorporated to help post process the raw results from visual inspection. This is beyond the scope of this paper but can be a challenging research topic for future work.
%
%
%
%

We also observe that the bottleneck for the whole pipeline is the RPN part. Although we spend a lot of efforts on the SPN classifier to achieve impressive classification accuracy, it is substantially degraded by the low detection recall rate for certain tiny components, such as test points, pins, or pads. As can be seen in Table \ref{tab:ablation}, there is a gap between CLF Acc and DET mAP. This may also originate from the issue of ambiguously  and subjectively defined bounding boxes, though it has long been a difficult research topic to detect objects of dramatic scale and aspect ratio variances.

\section{Conclusion}
In this paper, we present a dataset for generic component detection on PCBs and a model that can be optimized in a data-efficient way. Our method leverages low-shot similarity-based classification as well as a novel graph-based network to jointly leverage a small number of labeled instances from the test board and structure across the board to significantly improve classification. We successfully demonstrate the effectiveness of our proposed model on the challenging task of PCB component detection. We also examine the failed cases which result from the ambiguously and subjectively defined component bounding boxes and types. To resolve the ambiguity, in future work we plan to incorporate text information from the silk screen to post process the detection result. We also plan to better leverage structure in the region proposal network, which currently is a bottleneck as it does not explicitly deal with the domain shift present in our problem.

{\small
\bibliographystyle{ieee}
\bibliography{egbib}
}

\end{document}